\documentclass[journal]{IEEEtran}
\ifCLASSINFOpdf
\else
\fi
\usepackage[cmex10]{amsmath}
\usepackage{array}

\usepackage{mdwmath}
\usepackage{mdwtab}

\usepackage{eqparbox}
\usepackage{fixltx2e}
	\usepackage{url}

\usepackage{color}
\usepackage{psfrag}
\usepackage{graphicx}
\usepackage{wrapfig}
\usepackage{lscape}
\usepackage{rotating}
\usepackage{epstopdf}

\usepackage{graphicx}
\usepackage{epstopdf}
\usepackage{amssymb}
\usepackage{booktabs}
\usepackage{multirow}
\usepackage{multicol}
\usepackage[ruled,vlined]{algorithm2e}

\newcommand {\bOmega}{\boldsymbol{\Omega}}

\newcommand{\ncc}{\newcommand}
\ncc{\bs}{\boldsymbol}

\begin{document}
%
\title{Hyperspectral Subspace Identification Using SURE}
%
%
%

\author{Behnood~Rasti,~\IEEEmembership{Member,~IEEE,}
        ~Magnus~O.~Ulfarsson,~\IEEEmembership{Member,~IEEE,}
				 and Johannes~R.~Sveinsson,~\IEEEmembership{Senior Member,~IEEE,}
\IEEEcompsocitemizethanks{\IEEEcompsocthanksitem B. Rasti, M. O. Ulfarsson and J. R. Sveinsson are with the Department
of Electrical and Computer Engineering, University of Iceland, 107 Reykjavik, Iceland (corresponding author, e-mail: behnood.rasti@gmail.com)}}
\maketitle

\begin{abstract}
Remote sensing hyperspectral sensors collect large volumes of high dimensional
spectral and spatial data.  However, due to spectral and spatial redundancy
the true hyperspectral signal lies on a subspace of much lower dimension than
the original data.  The identification of the signal subspace is a very
important first step for most hyperspectral algorithms. In this paper we
investigate the important problem of identifying the hyperspectral signal subspace
by minimizing the mean squared error (MSE) between the true signal and an
estimate of the signal. Since the MSE is uncomputable in practice, due to its
dependency on the true signal, we propose a method based on the Stein's
unbiased risk estimator (SURE) that provides an unbiased estimate of the MSE.
The resulting method is simple and fully automatic and we evaluate it using both simulated
and real hyperspectral data sets.  Experimental results shows that our proposed method compares well to recent state-of-the-art subspace identification
methods.

\end{abstract}


%

\section{Introduction}
%
%
%
%
\IEEEPARstart{H}{yperspectral}
 remote sensing has the potential to detect and identify ground materials in remotely sensed scenes. As a result, it has found use in a great number of remote sensing applications.

Hyperspectral images (HSIs) contain detailed spectral information across a range of wavelengths. Thus, each pixel represents a spectral response. 
Having such a detailed spectral resolution decreases the spatial resolution \cite{varshney2010advanced}. Due to spectral redundancy, hyperspectral data lives in a low-rank subspace of the original high dimensional space (i.e., the rank of the subspace is much lower than the number of bands of the HSI). That fact is reflected in models typically used for hyperspectral data \cite{RastiSPIE13UWT,WSRRR,BourennaneTD}, where the unknown signal is modeled as a linear combination of few endmembers weighted by abundance fractions.


Since HSI generally lie on a low dimensional subspace the identification of it is a crucial task in HSI modeling and analysis.
An important step in the subspace identification is the problem of selecting the rank of the hyperspectral data and selecting it correctly is very important.
For instance, in unmixing application, over/under estimating the rank parameter can substantially change the result of endmember detection and consequently the corresponding abundance maps. Despite significant efforts the hyperspectral rank selection problem remains a big challenge. In the literature, signal subspace identification is also referred to by alternative names such as rank selection, intrinsic order selection, estimation of the number of endmembers, virtual dimension, estimation of number of spectrally distinct signal sources, etc. \cite{HySime,Keshava2003A,Changsubs}. In this paper, we use two terms, the subspace dimension and rank exchangeably.

 In \cite{HFC}, an eigenvalue based detector was developed by Harsanyi, Farrand and Chang called HFC. Later, HFC was extended to cope with the presence of non-uniform noise in the spectral direction of HSI, the resulting method was called NWHFC. Also, in \cite{AIC_MDL}, it was shown that information theoretic criteria such as the minimum description length (MDL) \cite{MDL} and Akaike's information criterion (AIC) \cite{AIC} are not suitable for identification of the signal subspace. Recently, a method for estimating spectral endmembers similar to HFC has been proposed in \cite{ELM}.

Hyperspectral subspace identification by minimum error (HySime) \cite{HySime} determines the subspace dimension (rank) by deriving an unbiased estimate of the MSE and is therefore similar to the method proposed in this paper. But, HySime is based on a different modeling for the hyperspectral data and thus leads to a different method. In \cite{Zare2007_RS}, an extension of iterated constrained endmember (ICE) \cite{ICE} was given by promoting sparsity prior to estimate the rank of hyperspectral data set. A rank selection approach was given in \cite{rarevectors} that uses the $\ell_{2,\infty}$ norm to select the subspace while accounting for the presence of rare signal. In \cite{MSE_Rank_sel}, the HSI rank was selected based on estimating the MSE using a simultaneous rank selection and denoising approach. Also, two geometry-based approaches (GENE-CH and GENE-AH) were proposed in \cite{GENE} for estimating the number of endmembers. GENE-CH and GENE-AH are based on the hypotheses that all the observed pixels lie in the convex hull and the affine hull of the endmember signatures, respectively. In addition, these algorithms rely on endmember estimation method called $p$-norm-based pure pixel identification (TRI-P).



Due to spatial redundancy, HSIs are sparse in a sparsifying dictionary such as wavelets
\cite{RastiB}. Wavelets have been widely used for sparse signal modeling since they are able to represent signals with sparse coefficients \cite{EladBook,Vetterli1995,Strang}. An integral part of sparse modeling is sparse estimation. Sparse estimation problems, such as matching pursuit (MP) \cite{MP}, least absolute shrinkage and selection operator (Lasso) \cite{Tibshirani94}, and basis pursuit denoising (BPDN) \cite{Chen98atomicdecomposition} have opened a broad research area called sparse signal processing or sparse signal analysis.
Sparse models are used in many HSI applications such as denoising, blind source separation and inpainting \cite{FORPjour,hypeGMCA,HyInpaint}. Here, we use a sparse model and estimation for hyperspectral subspace identification. 

Hyperspectral subspace dimension (rank) is a hyperspectral model parameter. 
Hence, we seek a model selection criteria for HSI which depends only on the observed data. 
A typical criterion used for model selection is the mean square error (MSE). The MSE is not computable since it requires knowledge of the true (unknown) signal. Therefore, computationally intensive methods such as cross-validation (CV) have been widely used for estimating the MSE, for model selection \cite{CVMS, modelCV} and tuning parameter selection \cite{CVSHURE, CVTV, CVTVRS}. 
Stein's unbiased risk estimator (SURE) is introduced in \cite{SURE} as an unbiased estimator of the MSE. SURE is used for selecting the thresholding parameter for wavelet shrinkage in \cite{Donoho95adaptingto} and then as a general method for selecting tuning parameter in \cite{SoloSURE}. Since then, SURE has been widely used for model and parameter selection in the area of signal processing \cite{CVSHURE,Eldar,mou} and see \cite{Magnus} for further references.

In this paper, we propose a hyperspectral subspace identification technique called hyperspectral SURE (HySURE). HySURE is a SURE-based technique which selects the subspace dimension and sparsity tuning parameters, simultaneously, for hyperspectral data. Previous subspace identification methods are only concerned about the spectral redundancy and therefore only focus on selecting the rank of the hyperspectral signal. Unlike previous methods HySURE also makes use of the spatial redundancy. In experiments, HySURE is compared with other techniques for both simulated and real HSI data sets. Also, it is shown that the sparse model used in this paper outperforms some other sparse models in terms of SURE value for real and simulated hyperspectral data.

The rest of the paper is organized as follows. After giving a short description of the notations used in the paper, a general sparse model and sparse estimation for HSI are presented in Section \ref{sec:HSImodelandEst} and \ref{sec:Estimation}, respectively. HySURE proposed as a hyperspectral rank selection method in Section \ref{sec:SURE}. The experimental results are given in Sections \ref{sec:SE} and \ref{sec:expReal} for simulated and real hyperspectral data sets, respectively. Finally, Section \ref{sec:con} concludes the paper.
\subsection{Notation}
\label{sec: Notation}

In this paper, the number of bands and pixels in each band are denoted by $p$ and $n$, respectively. Matrices are represented by bold and capital letters, column vectors by bold letters and the element placed in the $i$th row and $j$th column by $a_{ij}$. The identity matrix of size $p\times p$ is given by ${\bf I}_p$. $I$ stands for the indicator function and $\hat{\bf X}$ for the estimate of ${\bf X}$.
\section{Hyperspectral Image Modeling}
\label{sec:HSImodelandEst}

A hyperspectral image can be modeled by
\begin{equation*}
	{\bf Y} = {\bf X} + {\bf N},
\end{equation*}
where ${\bf Y}=\left[{\bf y}_{\left(i\right)}\right]$ is an $n \times p$ matrix containing the vectorized observed image at band $i$ in its $i$th columns, ${\bf X}=\left[{\bf x}_{\left(i\right)}\right]$ is the true unknown signal to be estimated and is represented as an $n \times p$ matrix containing the unknown vectorized image at band $i$ in its $i$th columns and ${\bf N}=\left[{\bf n}_{\left(i\right)}\right]$ is an $n \times p$ matrix containing the vectorized zero-mean Gaussian noise at band $i$ in its $i$th columns. Note that, the Gaussian noise model has been very commonly used in the remote sensing literature for hyperspectral modeling, e.g., \cite{HySime, GENE}. 
We assume that the noise is whitened by	the noise covariance matrix, ${\bf \Omega}=\mbox{diag}\left(\sigma^2_1,~\sigma^2_2,~\ldots,~\sigma^2_p\right)$ where $\sigma_p$ is the noise standard deviation in band $p$, thus after estimation the signal is restored by ${\hat{\bf X}}\bOmega^{1/2}$.

A general sparse model for the HSI is given by
\begin{equation}\label{eq: modelgen}
{\bf Y} = {\bf A}{\bf W}_r{\bf M}_r^T + {\bf N},
\end{equation}
where ${\bf A}=\left[{\bf a}_{\left(i\right)}\right]$ ($n\times n$ matrix) and ${\bf M}_r=\left[{\bf m}_{\left(i\right)}\right]$ ($p\times r$ matrix, $r \leq \mbox{min}\left(n,p\right)$) are known two dimensional and one dimensional orthogonal sparsifying bases, respectively, and ${\bf W}_r=\left[{\bf w}_{\left(i\right)}\right]$ ($n\times r$ matrix) contains unknown sparse coefficients for the unknown hyperspectral data, ${\bf X}$. In this paper we will focus on orthogonal wavelet bases. 
\section{Hyperspectral Image Estimation}
\label{sec:Estimation}

Penalized least squares has been widely used for estimating signal processing models. Penalized least squares is the sum of a squared error term and a penalty. The penalty term should be chosen based on the prior knowledge of the signal. Here, the $\ell_1$ penalty is used to promote sparsity. Since remote sensing signals such as HSIs are not inherently sparse, they need to be expanded in a basis where they have sparse representation. 
The sparse estimation problem for the general model (\ref{eq: modelgen}) is given by
\begin{equation}\label{eq: cost}
	{\hat {{\bf W}}}_r=\arg\min_{{\bf W}_r}~\frac{1}{2}\left\|{\bf Y}-{\bf A}{\bf W}_r{\bf M}_r^T\right\|^{2}_{F}+\sum_{t,k}\lambda\left|w_{tk}\right|.
\end{equation}
It can be seen that the solution to (\ref{eq: cost}) is given by the following shrinkage function (see Appendix \ref{App: Shrinkage})
\begin{equation}\label{eq: Sol}
\hat{w}_{tk}=\max\left(0,\left|b_{tk}\right|-\lambda\right) \frac{b_{tk}}{\left|b_{tk}\right|}.
\end{equation}
where ${\bf B}={\bf A}^T{\bf Y}{\bf M}_r=\left[b_{tk}\right]$.
%
%
%
%
\section{Estimating MSE Using SURE}
\label{sec:SURE}

In (\ref{eq: cost}), $\lambda$ and $r$ are unknown and we want to select them so that they minimize the MSE,
\begin{equation}\label{Risk}
 R_{{ \lambda},r}=E\left\|{\bf X}-\hat{\bf X}_{{ \lambda},r}\right\|_F^2,
\end{equation}
where ${ \lambda}$ is the sparsity tuning parameter and $r$ is the rank number. 
 Unfortunately, in most applications such as HSI the true signal ${\bf X}$ is unknown and thus it is impossible to compute the MSE. 
For deterministic signals in Gaussian noise such as (\ref{eq: modelgen}) an unbiased estimator of the MSE called SURE can be derived. The general form of SURE is given by
\begin{equation*}
{\hat R}_{{ \lambda},r} =\left\|{\bf E}\right\|_F^2+2\sum_{j=1}^p \mbox{tr}\left( \bs{\Omega}  \frac{\partial{\hat{\bf x}}_{(j)}}{\partial{\bf y}_{(j)}^T}\right)-np,
\end{equation*}
where ${\bf E}={\bf Y}-{\bf A}{\hat{\bf W}}_r{\bf M}_r^T$ is the residual.  In Appendix B it is shown that SURE for (\ref{eq: cost}) is given by
\begin{equation}
{\hat R}_{{ \lambda},r}= \left\|{\bf E}\right\|_F^2+2\mbox{ed}(r,\lambda)-np \label{HySURE}
\end{equation}
where $\mbox{ed}(r,\lambda)$ is the effective dimensionality of the subspace identified and is given by
\begin{eqnarray}
\mbox{ed}(r,\lambda)=\sum_{t=1}^n\sum_{k=1}^r{I}(\left|b_{tk}\right|>\lambda). \label{ed}
\end{eqnarray}
From (\ref{ed}) it is clear that $\mbox{ed}(r,\lambda)$ is simply the number of coefficients in the subspace identified determined by the rank $r$ and the number of
coefficients in an orthogonal sparsifying basis, such as wavelets, that survive the threshold $\lambda$.  The SURE formula (\ref{HySURE}) makes it clear that the method is a tradeoff between the fit of the model $\|\bf E\|_F^2$ and the effective dimensionality $\mbox{ed}$.  The fit generally improves ($\| \bf E\|^2_F$ decreases) with increasing $r$ and decreasing $\lambda$, but on the other hand $\mbox{ed}$ increases. So one expects that SURE takes a minimum for some $0<r<p$ and $\lambda>0$.

\begin{figure}[tbp]
\centering
\includegraphics[width=8.5cm]{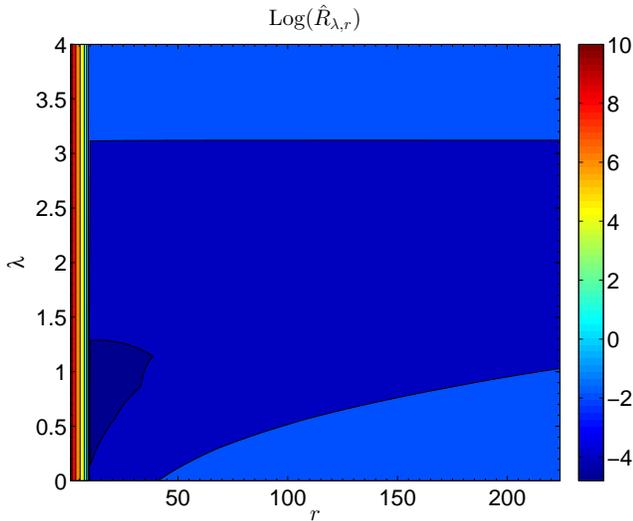}

		\caption{Contour plot in a logarithmic scale of SURE w.r.t. the rank number, $r$, and the sparsity tuning parameter, $\lambda$, for simulated hyperspectral data set 1.}
	\label{fig:Synthetic_sure_mse_lr_rs}
\end{figure}

To use SURE in practice a grid is selected for $\lambda$ and $r$ and then SURE is computed for each grid point. The optimal rank and tuning parameter $\lambda$ are selected based on the minimum of SURE:
\begin{equation*}
(\hat{\lambda},\hat{r}) = \arg\min_{\lambda,r}\hat{R}_{\lambda,r}.
\end{equation*}
As an example on the use of SURE we apply it on simulated data set 1 (described below) having SNR=15 dB. Fig. \ref{fig:Synthetic_sure_mse_lr_rs} depicts how SURE selects model parameters, the rank and the sparsity tuning parameters, simultaneously.  Fig. \ref{fig:Synthetic_sure_mse_lr_rs} is a contour plot of the risk value in a logarithmic scale for $1\leq r\leq 224$ and $0\leq\lambda\leq 4$. SURE selects $r=8$ and $\lambda=0.44$ for this experiment, since they give the lowest risk value.

\subsection{Hyperspectral SURE}
The model presented in (\ref{eq: modelgen}) allows for some flexibility, i.e. $\bf {A}$ and $\bf {M}_r$ can be selected in various ways.  In this paper we select $\bf A$ as the two dimensional orthogonal wavelet basis and $\bf {M}_r$ as a low rank matrix of spectral eigenvectors.  We will justify this choice in the experimental section.  We call the method that results from this choice hyperspectral SURE (HySURE).

\section{Simulations and Evaluations}
\label{sec:SE}



As we mentioned above model {(\ref{eq: modelgen}) allows for some flexibility in the selection of $\bf{A}$ and $\bf{M}_r$ and the HySURE is the method that follows when the quantities are selected as 2D wavelet basis and a low rank spectral eigenvector matrix, respectively.  Here we justify the choice of HySURE by comparing it with six alternative models (all in the form of the general model) given in Table \ref{tab:Spect}. For convenience, all the models are numbered in Table \ref{tab:Spect} with brief descriptions so we call them by numbers (1-7) in the rest of the paper.  Note that HySURE uses model (7). Also, in this section, HySURE is compared with four other HSI rank selection techniques (NWHFC, HySime, GENE-CH and GENE-AH) from the literature in several experiments by using two simulated hyperspectral data sets. In experiments, five level Daubechies wavelet with eight coefficients for the spatial basis and with two coefficients for the spectral basis are used.
\begin{table*}[htbp]
  \addtolength{\tabcolsep}{5pt}
	\centering
  \caption{Seven linear sparse HSI models used in the model selection experiments.}
 \begin{tabular}{||c|l|c|c|c||}
		\hline\hline
    & HSI Model formulation & Spatial basis used & Spectral basis used  & Spectral Rank \\ \hline\hline
Model (1)	   & ${\bf Y} = {\bf D}_2{\bf W}{\bf D}_1^T + {\bf N}$ & 2-D wavelet & 1-D wavelet & Full-Rank\\ \hline
Model (2)	   & ${\bf Y} = {\bf D}_2{\bf W} + {\bf N}$& 2-D wavelet & No spectral basis & Full-Rank\\ \hline
Model (3)	   & ${\bf Y} = {\bf W}{\bf D}_1^T + {\bf N}$ & No spatial basis & 1-D wavelet & Full-Rank\\ \hline
Model (4)	   & ${\bf Y} = {\bf W}{\bf V}^T + {\bf N}$ & No spatial basis & Spectral eigenvectors & Full-Rank\\ \hline
Model (5)	   & ${\bf Y} = {\bf D}_2{\bf W}{\bf V}^T + {\bf N}$ & 2-D wavelet & Spectral eigenvectors & Full-Rank\\ \hline
Model (6)	   & ${\bf Y} = {\bf W}{\bf V}_r^T + {\bf N}$&No spatial basis & Spectral eigenvectors & Low-Rank\\ \hline
Model (7)	   & ${\bf Y} = {\bf D}_2{\bf W}{\bf V}_r^T + {\bf N}$ & 2-D wavelet & Spectral eigenvectors & Low-Rank \\
		\hline
		\hline
    \end{tabular}%
  \label{tab:Spect}%
\end{table*}%



\subsection{Hyperspectral Image Simulation}
As previously mentioned, computing MSE for a real data is impossible since the true signal is unknown. Therefore, in this paper simulated hyperspectral data sets are used so we are able to compute the MSE and compare it with SURE. The comparison based on the simulated data shows the reliability of SURE as an estimator for MSE. Additionally, the proposed HSI rank selection technique is evaluated based on two simulated data sets and compared with other rank selection techniques.

\subsubsection{Data set 1}
The first hyperspectral data set is simulated by randomly selecting 10 different endmembers from the USGS spectral library \cite{USGS} shown in Fig. \ref{Endmembers}. The associated abundance fractions are shown in Fig. \ref{abundances}. A linear mixture model is used to create hyperspectral image of size $128\times128\times224$ where the Gaussian noise is allowed to be band-dependent with the noise variance at band $i$ given by \cite{HySime} 
\begin{equation}\nonumber
	 \sigma_i^2=\sigma^2\frac{e^{-\frac{\left(i-p/2\right)^2}{2\eta^2}}}{\sum^p_{j=1} e^{-\frac{\left(j-p/2\right)^2}{2\eta^2}}}.
\end{equation}
 Band 112 of simulated hyperspectral data set 1, both before and after adding noise is given in Fig. \ref{fig:band112_noise_spec}. Also, in Fig. \ref{fig:band112_noise_spec}, the spectrum located at position (64,64) is depicted before and after adding noise. For this data set $\eta=1/18$. In the experiments, this data set is used for both model selection and rank selection.  Note that, in simulated data set 1, spatial structure and pixel variability have been taken into account for simulating fractional abundances so we get closer simulation compared to the real case which makes rank estimation more challenging. However, in experiments, we will also consider the case of applying sum to one assumption on the fractional abundances.
\begin{figure}[htbp]
	\centering
		\includegraphics[width=8.5cm]{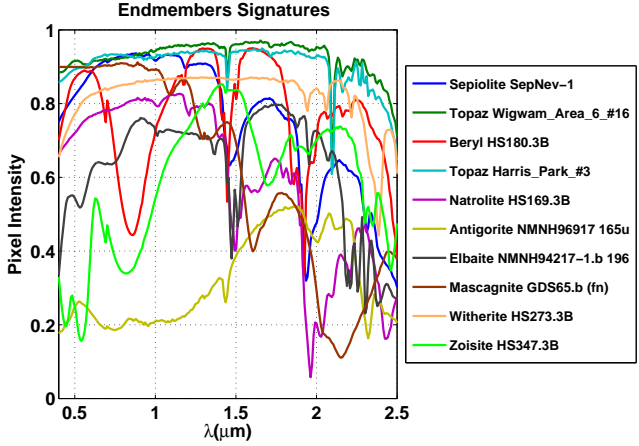}
		\caption{Ten endmembers signatures from USGS library used for simulated hyperspectral data set 1.}
	\label{Endmembers}
\end{figure}
\begin{figure*}[htbp]
	\centering
		\includegraphics[width=18cm]{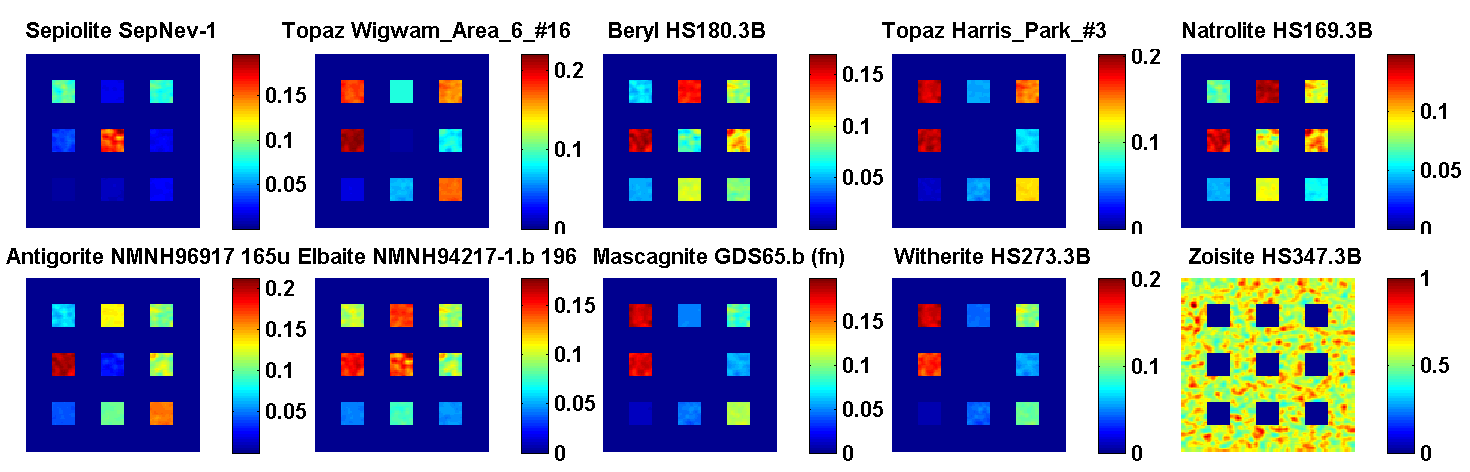}
		\caption{Fractional abundances associated to ten endmembers selected from USGS library and used for simulated hyperspectral data set 1.}
	\label{abundances}
\end{figure*}
\begin{figure}[htbp]
	\centering
		\includegraphics[width=9cm]{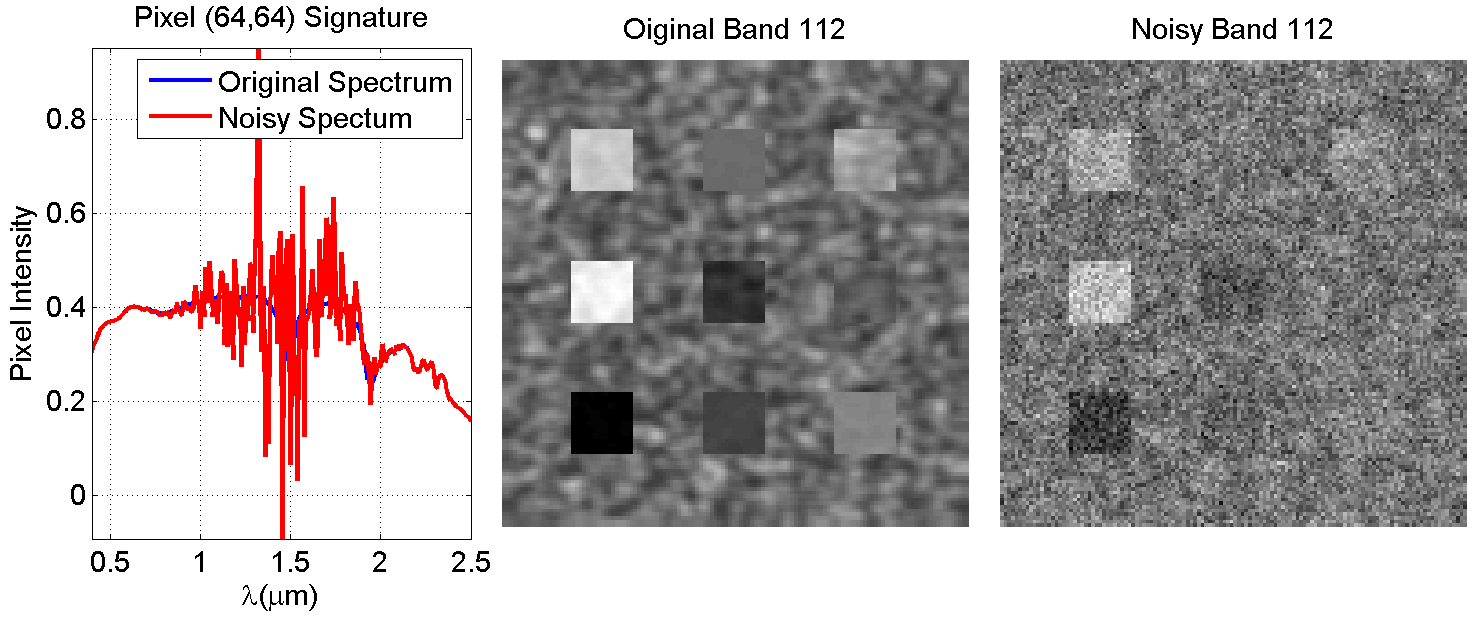}
	\caption{Band 112 and the spectrum located at position (64,64) of simulated hyperspectral data set 1 before and after adding noise (SNR=15 dB).}
	\label{fig:band112_noise_spec}
\end{figure}

\subsubsection{Data set 2}

The second simulated hyperspectral data set is based on a linear mixture model having the same size as data set 1. Endmembers are also selected randomly from the USGS spectral library. However, the fractional abundances are generated based on the Dirichlet distribution \cite{HySime,GENE}. Noisy data set is simulated as previously explained for data set 1 for two different cases, $\eta=1/18$ and $\eta=0$. The Matlab code to generate data set 2 can be found online \cite{HySimeCode}. In this paper, this data set is only used for evaluating the rank selection techniques.

\subsection{Hyperspectral Model Selection Using SURE}
In this subsection, we use SURE to compare the seven aforementioned models (1-7) and select the one which gives the lowest risk value for HSI modeling. We should note that for HSI model evaluation we do not need a simulated data set since SURE can be directly applied on real data set (see Section \ref{subsec: HSIMSReal}). Here, simulated data set 1 (SNR=15 dB) is used to investigate how well SURE estimates the MSE. 

Fig. \ref{fig:Synthetic_sure_mse} compares the MSE and SURE values for models (1-5)  as a function of the tuning parameter, $\lambda$. It can be seen that SURE (dash lines) is a good estimator of MSE (solid lines) for all the five models. Also, in Fig. \ref{fig:Synthetic_sure_mse}, it can be seen that model (3), where the hyperspectral data is only projected spectrally on wavelet basis, has the worst performance and model (5) outperforms other models based on SURE.

Among the purely wavelet based models (1), (2) and (3) the spectral-spatial model (1) is the best, followed by the spatial model (2), and trailed by the spectral model (3).

Fig. \ref{fig:Synthetic_sure_mse} also shows that model (4) in which the hyperspectral image is projected on the spectral eigenvectors (${\bf V}$), is a better model than the spatial-spectral wavelet model (1) since it has a lower minimum SURE value. 

Overall, model (5) has the lowest SURE value which demonstrates that projecting the simulated hyperspectral image on the wavelet basis spatially and on its spectral eigenvectors (${\bf V}$) spectrally has the lowest SURE value. That is better seen in Fig. \ref{fig:Synthetic_sure_mse_lr_ps}.
\begin{figure}[htbp]
	\centering
		\includegraphics[width=8.5cm]{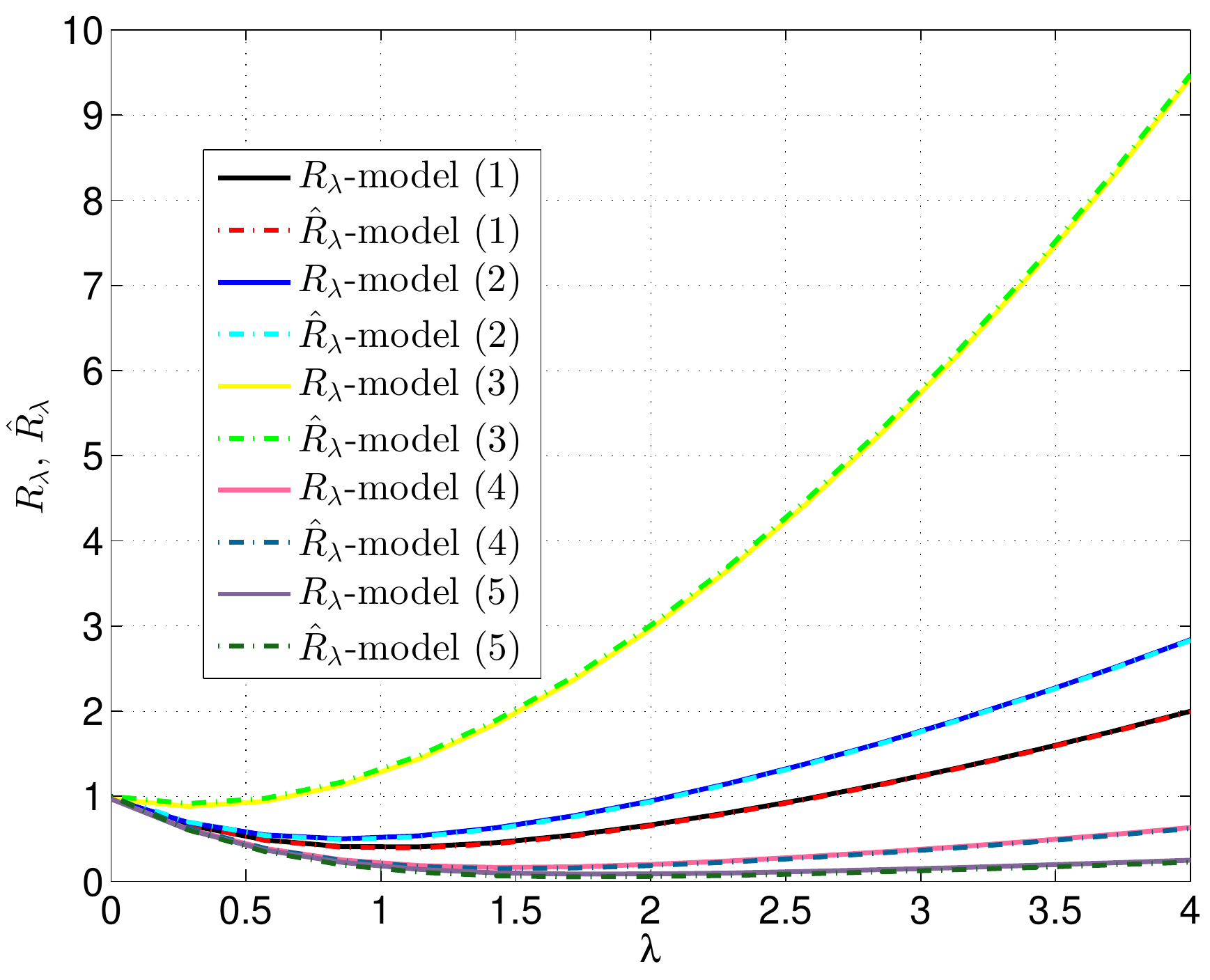}
		\caption{A comparison of MSE with SURE for models (1-5) w.r.t. the tuning parameter $\lambda$ for simulated hyperspectral data set 1.}
	\label{fig:Synthetic_sure_mse}
\end{figure}


%

Fig. \ref{fig:Synthetic_sure_mse_lr_ps} shows the low-rank models (6) and (7) for $r=10$, w.r.t. the tuning parameter, $\lambda$. It can be seen that, also for low-rank models SURE estimates the MSE successfully. From Fig. \ref{fig:Synthetic_sure_mse_lr_ps}, it can also be seen that model (7) outperforms model (6) in terms of the risk value. 

SURE and MSE for full-rank models (4) and (5) are also displayed in Fig. \ref{fig:Synthetic_sure_mse_lr_ps}. It can be seen that, when $\lambda$ is small the low-rank models substantially outperform the full-rank ones but when $\lambda$ increases, the full-rank and low-rank models perform similarly. This can be explained by the fact that for low $\lambda$ the low energy (noisy) components are neglected by the low-rank models but kept by the full-rank models, but for high $\lambda$ the noisy components are zeroed out due to high threshold value which makes the low-rank and full-rank models perform similarly. 
By comparing results in Fig. \ref{fig:Synthetic_sure_mse} and \ref{fig:Synthetic_sure_mse_lr_ps} one can conclude that model (7) gives the lowest risk value for all the seven models for the simulated HSI. 
 Note that, in this experiment, for low-rank models the true rank number is selected i.e. $r=10$.
\begin{figure}[htbp]
	\centering
		\includegraphics[width=8.5cm]{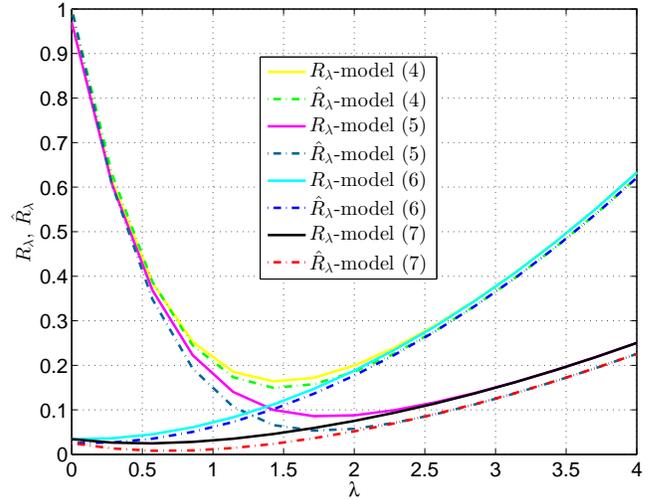}
		\caption{A comparison of MSE with SURE for models (4-7) w.r.t. the tuning parameter $\lambda$ for simulated hyperspectral data set 1.}
	\label{fig:Synthetic_sure_mse_lr_ps}
\end{figure}

\subsection{Evaluation of HySURE}

Here, we compare HySURE with NWHFC, HySime, GENE-CH and GENE-AH based on two simulated hyperspectral data sets. In this paper, the noise is estimated by the multiple regression technique given in \cite{HySime} for all techniques including the noise whitening step for NWHFC and therefore the comparisons of the aforementioned rank selection techniques will not be dependent on the estimation of noise. The false alarm parameter $P_{FA}$ is selected to be $P_{FA}=10^{-8}$ for NWHFC and the geometric methods (GENE-CH and GENE-AH) and also $p=2$ for TRI-P algorithm used in the geometric techniques as suggested in \cite{GENE}.

\subsubsection{Data set 1; Experiment 1}
In Table \ref{tab:rank_sel_syn}, we compare HySURE with NWHFC, HySime, GENE-CH and GENE-AH, for simulated data set 1. The methods are evaluated for different levels of SNR (10, 15, 20, 25, 35 and 50 dB) and different rank numbers, $3\leq r\leq 10$. The results are obtained by taking the median over 10 experiments. In this experiment, for the geometric methods (GENE-CH and GENE-AH) maximum rank number ($r_{max}$) is chosen to be 25 since the maximum rank number of the simulated data set used in the experiment is 10. Obviously, the rank estimation problem becomes more challenging by decreasing SNR and also with increasing the number of endmembers since the complexity of the mixing problem is increased. As can be seen in Table \ref{tab:rank_sel_syn}, with increasing $r$ and decreasing SNR, HySime underestimates the rank number, GENE-CH and GENE-AH fail to estimate the true rank in many cases and NWHFC fails for all the cases except $r=3$. HySURE performs successfuly for SNR= 50 dB and 35 dB. It estimates the true rank for $r\leq 8$ in the case of SNR=25 and 20 dB. Also, in the case of low SNRs, SNR=15 and 10 dB, HySURE estimates the true rank for $r \leq 6$ and $r \leq 5$, respectively. We should note that for the cases that HySURE fails, the estimations are very close to the true ranks. 
\begin{table}[htbp]
  \centering\addtolength{\tabcolsep}{-1.5pt}
  \caption{Hyperspectral rank selection using five techniques (HySURE, NWHFC, HySime, GENE-CH, GENE-AH) applied on simulated data set 1 for SNR= 10, 15, 20, 25, 35 and 50, in decibel and $3\leq r\leq 10$. The correct rank estimations are shown by bold numbers. The results are obtained by taking the median over 10 experiments.}
    \begin{tabular}{c|l|cccccccc}
    \toprule
    SNR & Method & r=3   & r=4   & r=5   & r=6   & r=7   & r=8   & r=9   & r=10 \\
    \midrule
          & HySURE & \textbf{3} & \textbf{4} & \textbf{5} & \textbf{6} & \textbf{7} & \textbf{8} & \textbf{9} & \textbf{10} \\
          & NWHFC & \textbf{3} & 3     & 2     & 3     & 2     & 2     & 4     & 4 \\
    50 dB & HySime & \textbf{3} & \textbf{4} & 4     & 5     & 5     & 6     & 7     & 7 \\
          & GENE-CH & 8     & 6     & 9     & 7     & 11    & 13    & 13    & 17 \\
          & GENE-AH & 4     & 6     & \textbf{5} & 7     & \textbf{7} & 9     & 11    & 10 \\
					\midrule
          & HySURE & \textbf{3} & \textbf{4} & \textbf{5} & \textbf{6} & \textbf{7} & \textbf{8} & \textbf{9} & \textbf{10} \\
          & NWHFC & \textbf{3} & 3     & 2     & 3     & 2     & 3     & 3     & 4 \\
    35 dB & HySime & \textbf{3} & 3     & 4     & 4     & 4     & 5     & 5     & 5 \\
          & GENE-CH & 5     & 10    & 6     & \textbf{6} & 14    & 6     & 10    & 12 \\
          & GENE-AH & 5     & 7     & 6     & \textbf{6} & 5     & 7     & 6     & 6 \\
					\midrule
          & HySURE & \textbf{3} & \textbf{4} & \textbf{5} & \textbf{6} & \textbf{7} & \textbf{8} & 8     & 9 \\
          & NWHFC & \textbf{3} & 3     & 2     & 3     & 2     & 3     & 2     & 4 \\
    25 dB & HySime & \textbf{3} & 3     & 4     & 4     & 4     & 4     & 4     & 4 \\
          & GENE-CH & 5     & 5     & 6     & 5     & 6     & 5     & 9     & 7 \\
          & GENE-AH & 6     & 5     & \textbf{5} & \textbf{6} & 6     & 5     & 11    & 5 \\
					\midrule
          & HySURE & \textbf{3} & \textbf{4} & \textbf{5} & \textbf{6} & \textbf{7} & \textbf{8} & 8     & 9 \\
          & NWHFC & \textbf{3} & 3     & 2     & 3     & 2     & 3     & 2     & 4 \\
    20 dB & HySime & \textbf{3} & 3     & 3     & 4     & 3     & 3     & 4     & 3 \\
          & GENE-CH & 6     & 3     & 3     & 4     & 5     & 3     & 11    & 8 \\
          & GENE-AH & 6     & 3     & 3     & 8     & 5     & 3     & 6     & 4 \\
					\midrule
          & HySURE & \textbf{3} & \textbf{4} & \textbf{5} & \textbf{6} & 6     & 7     & 8     & 9 \\
          & NWHFC & \textbf{3} & 3     & 2     & 3     & 3     & 3     & 2     & 3 \\
    15 dB & HySime & 2     & 2     & 2     & 3     & 3     & 2     & 2     & 2 \\
          & GENE-CH & \textbf{3} & 3     & 4     & 3     & 3     & 4     & 8     & 4 \\
          & GENE-AH & \textbf{3} & 3     & 4     & 3     & 3     & 4     & 8     & 3 \\
					\midrule
          & HySURE & \textbf{3} & \textbf{4} & \textbf{5} & 5     & 6     & 7     & 8     & 8 \\
          & NWHFC & \textbf{3} & 3     & 2     & 3     & 3     & 3     & 2     & 3 \\
     10 dB & HySime & 2     & 1     & 2     & 2     & 2     & 2     & 2     & 2 \\
          & GENE-CH & \textbf{3} & 3     & 4     & 3     & 3     & 3     & 4     & 3 \\
          & GENE-AH & \textbf{3} & 3     & 4     & 3     & 3     & 3     & 4     & 3 \\

    \bottomrule
    \end{tabular}%
  \label{tab:rank_sel_syn}%
\end{table}%

\subsubsection{Data set 1; Experiment 2}

Here, we repeat the rank selection experiment on data set 1 after applying sum to one assumption on the abundances. The rank estimation results are given in Table \ref{tab:rank_sel_syn2}. It can be seen that NWHFC gives better results compared to the previous experiment. Although NWHFC fails in many cases, the estimations are closer to the true ranks in this experiment. It appears that there are no trends in the behaviors of GENE-CH and GENE-AH and like in the previous experiments these methods fail in many cases. The estimation results for HySime and HySURE show slight improvements from the previous experiment. For instance, HySime can estimate the true rank for $r=3$, and also for $r=4$ when SNR$\geq 35$. Overall, HySURE considerably outperforms the other rank selection techniques based on the experiments carried out on simulated data set 1.

\begin{table}[htbp]
  \centering\addtolength{\tabcolsep}{-1.5pt}
  \caption{Hyperspectral rank selection using five techniques (HySURE, NWHFC, HySime, GENE-CH, GENE-AH) applied on simulated data set 1 having sum to one constraint on fractional abundances for SNR= 10, 15, 20, 25, 35 and 50, in decibel and $3\leq r\leq 10$. The correct rank estimations are shown by bold numbers. The results are obtained by taking the median over 10 experiments.}
    \begin{tabular}{c|l|cccccccc}
    \toprule
    SNR & Method & r=3   & r=4   & r=5   & r=6   & r=7   & r=8   & r=9   & r=10 \\
    \midrule
          & HySURE & \textbf{3} & \textbf{4} & \textbf{5} & \textbf{6} & \textbf{7} & \textbf{8} & \textbf{9} & \textbf{10} \\
          & NWHFC & 6     & 9     & 7     & \textbf{6} & \textbf{7} & \textbf{8} & \textbf{9} & \textbf{10} \\
    50 dB & HySime & \textbf{3} & \textbf{4} & 4     & 5     & 5     & 6     & 7     & 7 \\
          & GENE-CH & \textbf{3} & \textbf{4} & \textbf{5} & 9     & 5     & 6     & 6     & 12 \\
          & GENE-AH & \textbf{3} & \textbf{4} & \textbf{5} & 5     & 5     & 6     & \textbf{9} & 7 \\
					\midrule
          & HySURE & \textbf{3} & \textbf{4} & \textbf{5} & \textbf{6} & \textbf{7} & \textbf{8} & \textbf{9} & \textbf{10} \\
          & NWHFC & 4     & 5     & 6     & \textbf{6} & \textbf{7} & \textbf{8} & \textbf{9} & \textbf{10} \\
    35 dB & HySime & \textbf{3} & \textbf{4} & 4     & 4     & 5     & 4     & 4     & 4 \\
          & GENE-CH & 5     & \textbf{4} & 6     & 5     & 6     & 6     & 6     & 6 \\
          & GENE-AH & 4     & \textbf{4} & 6     & 5     & 6     & 6     & 6     & 6 \\
					\midrule
          & HySURE & \textbf{3} & \textbf{4} & \textbf{5} & \textbf{6} & \textbf{7} & \textbf{8} & \textbf{9} & 9 \\
          & NWHFC & \textbf{3} & 5     & 6     & \textbf{6} & \textbf{7} & \textbf{8} & 7     & 9 \\
    25 dB & HySime & \textbf{3} & 3     & 4     & 4     & 4     & 4     & 4     & 4 \\
          & GENE-CH & 7     & 6     & 7     & 7     & \textbf{7} & 6     & 6     & 5 \\
          & GENE-AH & 7     & 6     & 7     & 5     & 6     & 6     & 6     & 5 \\
					\midrule
          & HySURE & \textbf{3} & \textbf{4} & \textbf{5} & \textbf{6} & \textbf{7} & \textbf{8} & 8     & 9 \\
          & NWHFC & \textbf{3} & 5     & 6     & \textbf{6} & \textbf{7} & \textbf{8} & 7     & 9 \\
    20 dB & HySime & \textbf{3} & 3     & 4     & 4     & 4     & 3     & 4     & 3 \\
          & GENE-CH & 8     & 8     & \textbf{5} & 5     & \textbf{7} & 4     & 5     & 9 \\
          & GENE-AH & 6     & 8     & \textbf{5} & 5     & \textbf{7} & 4     & 5     & 6 \\
					\midrule
          & HySURE & \textbf{3} & \textbf{4} & \textbf{5} & \textbf{6} & \textbf{7} & 7     & 8     & 9 \\
          & NWHFC & \textbf{3} & 5     & \textbf{5} & \textbf{6} & 6     & 7     & 7     & 9 \\
    15 dB & HySime & \textbf{3} & 3     & 3     & 4     & 3     & 2     & 2     & 2 \\
          & GENE-CH & 14    & 7     & \textbf{5} & 5     & \textbf{7} & \textbf{8} & 4     & 5 \\
          & GENE-AH & 10    & 7     & \textbf{5} & 5     & \textbf{7} & \textbf{8} & 4     & 5 \\
					\midrule
          & HySURE & \textbf{3} & \textbf{4} & \textbf{5} & \textbf{6} & 6     & 7     & 8     & 9 \\
          & NWHFC & \textbf{3} & 5     & \textbf{5} & \textbf{6} & 6     & 7     & 7     & 9 \\
     10 dB & HySime & \textbf{3} & 3     & 2     & 3     & 3     & 2     & 2     & 2 \\
          & GENE-CH & 7     & 6     & 5     & 5     & 6     & \textbf{8} & 3     & 5 \\
          & GENE-AH & 7     & 6     & 5     & 5     & 6     & 6     & 3     & 5 \\

    \bottomrule
    \end{tabular}%
 \label{tab:rank_sel_syn2}%
\end{table}%

\subsubsection{Data set 2; Experiment 1}

HySURE is also compared with NWHFC, HySime, GENE-CH and GENE-AH based on simulated data set 2 for $\eta=1/18$. Table \ref{tab:rank_sel_syn_Data2} gives the results obtained by rank selection techniques for $r$ = 3, 5, 10, 15, 20 and 30 and SNR = 10, 15, 20, 25, 35 and 50. The results are obtained by taking the median over 10 experiments. 
All the parameters are selected like in the previous experiment except here $r_{max}=50$ (for GENE-CH and GENE-AH) since the maximum rank number used in this experiment is 30.

In this experiment, HySURE behaves as a robust rank selection technique w.r.t. the rank number and also to the noise power. HySURE estimates the true rank for all the cases. It appears that increasing noise power and number of endmembers affect HySime estimations. We should note that HySime is closely related to HySURE, but HySURE is based on sparse model and sparse estimation technique for the hyperspectral signal which makes it be more robust to the noise power. NWHFC and the geometric techniques does not perform satisfactorily and GENE-CH fails to select a rank in many cases where given by zeros in the table. Note that HySime and HySURE are parameter free techniques which is 	important for big data analysis.

\begin{table}[htbp]
  \centering
    \caption{Hyperspectral rank selection using five techniques (HySURE, NWHFC, HySime, GENE-CH and GENE-AH) applied on simulated data set 2 when $\eta=1/18$, for SNR= 10, 15, 20, 25, 35 and 50, in decibel and $r$= 3, 5, 10, 15, 20 and 30. The correct rank estimations are shown by bold numbers. The results are obtained by taking the median over 10 experiments.}
    \begin{tabular}{c|l|cccccc}
    \toprule
    SNR & Method & r=3   & r=5   & r=10  & r=15  & r=20  & r=30 \\
    \midrule

          & HySURE & \textbf{3} & \textbf{5} & \textbf{10} & \textbf{15} & \textbf{20} & \textbf{30} \\
          & NWHFC & 16    & 33    & 18    & 30    & 24    & 25 \\
    50 dB & HySime & \textbf{3} & \textbf{5} & \textbf{10} & \textbf{15} & 19    & 28 \\
          & GENE-CH & \textbf{3} & \textbf{5} & 11    & 19    & 29    & 0 \\
          & GENE-AH & \textbf{3} & \textbf{5} & \textbf{10} & 16    & 22    & 34 \\
					\midrule
          & HySURE & \textbf{3} & \textbf{5} & \textbf{10} & \textbf{15} & \textbf{20} & \textbf{30} \\
          & NWHFC & 7     & 21    & 9     & 20    & 10    & 15 \\
    35 dB & HySime & \textbf{3} & \textbf{5} & \textbf{10} & 14    & 18    & 24 \\
          & GENE-CH & \textbf{3} & 6     & 15    & 0     & 0     & 0 \\
          & GENE-AH & \textbf{3} & 6     & 12    & 24    & 33    & 45 \\
					\midrule
          & HySURE & \textbf{3} & \textbf{5} & \textbf{10} & \textbf{15} & \textbf{20} & \textbf{30} \\
          & NWHFC & 8     & 13    & 7     & 16    & 9     & 12 \\
    25 dB & HySime & \textbf{3} & \textbf{5} & \textbf{10} & 14    & 13    & 18 \\
          & GENE-CH & \textbf{3} & 7     & 26    & 0     & 0     & 0 \\
          & GENE-AH & \textbf{3} & 9     & 20    & 34    & 38    & 47 \\
					\midrule
          & HySURE & \textbf{3} & \textbf{5} & \textbf{10} & \textbf{15} & \textbf{20} & \textbf{30} \\
          & NWHFC & 5     & 10    & 7     & 14    & 7     & 11 \\
    20 dB & HySime & \textbf{3} & \textbf{5} & 9     & 11    & 10    & 14 \\
          & GENE-CH & \textbf{3} & 8     & 27    & 0     & 0     & 0 \\
          & GENE-AH & \textbf{3} & 8     & 19    & 37    & 30    & 46 \\
					\midrule
          & HySURE & \textbf{3} & \textbf{5} & \textbf{10} & 15    & \textbf{20} & \textbf{30} \\
          & NWHFC & 5     & 9     & 7     & 12    & 6     & 12 \\
    15 dB & HySime & \textbf{3} & \textbf{5} & 6     & 9     & 7     & 9 \\
          & GENE-CH & 4     & 9     & 32    & 0     & 40    & 0 \\
          & GENE-AH & 4     & 7     & 20    & 38    & 29    & 36 \\
					\midrule
          & HySURE & \textbf{3} & \textbf{5} & \textbf{10} & \textbf{15} & \textbf{20} & \textbf{30} \\
          & NWHFC & \textbf{3} & 8     & 7     & 11    & 5     & 11 \\
     10 dB & HySime & \textbf{3} & \textbf{5} & 5     & 7     & 6     & 7 \\
          & GENE-CH & 6     & 8     & 22    & 0     & 27    & 0 \\
          & GENE-AH & 5     & 7     & 21    & 38    & 30    & 27 \\

    \bottomrule
    \end{tabular}%
  \label{tab:rank_sel_syn_Data2}%
\end{table}%

\subsubsection{Data set 2; Experiment 2}

In this experiment, we add Gaussian noise when $\eta=0$ to the simulated data set 2. All rank selection techniques are compared in Table \ref{tab:rank_sel_syn_Data2_E2} for $r$ = 3, 5, 10, 15, 20 and 30 and SNR = 10, 15, 20, 25, 35 and 50. All parameters are selected like in the previous experiment. The results are obtained by taking the median over 10 runs. 
From Table \ref{tab:rank_sel_syn_Data2_E2}, it can be seen that all techniques estimate the true rank for $r=3$ for all SNR values. However, NWHFC fails for all the other cases shown in the table. Also, in Table \ref{tab:rank_sel_syn_Data2_E2}, we can see that the other four techniques (HySURE, HySime, GENE-CH and GENE-AH) perform similarly. Also, GENE-CH slightly outperform the other techniques in this experiment. Overall, by increasing the rank number and decreasing the signal to noise ratio, the rank estimation techniques underestimate the true rank.

\begin{table}[htbp]
  \centering
    \caption{Hyperspectral rank selection using five techniques (HySURE, NWHFC, HySime, GENE-CH and GENE-AH) applied on simulated data set 2 when $\eta=0$, for SNR= 10, 15, 20, 25, 35 and 50, in decibel and $r$= 3, 5, 10, 15, 20 and 30. The correct rank estimations are shown by bold numbers. The results are obtained by taking the median over 10 experiments.}
    \begin{tabular}{c|l|cccccc}
    \toprule
    SNR & Method & r=3   & r=5   & r=10  & r=15  & r=20  & r=30 \\
    \midrule
          & HySURE & \textbf{3} & \textbf{5} & \textbf{10} & \textbf{15} & \textbf{20} & 29 \\
          & NWHFC & \textbf{3} & 4     & 5     & 7     & 7     & 6 \\
    50 dB & HySime & \textbf{3} & \textbf{5} & \textbf{10} & 15    & 19    & 28 \\
          & GENE-CH & \textbf{3} & \textbf{5} & \textbf{10} & \textbf{15} & \textbf{20} & \textbf{30} \\
          & GENE-AH & \textbf{3} & \textbf{5} & \textbf{10} & \textbf{15} & \textbf{20} & \textbf{30} \\
					\midrule
          & HySURE & \textbf{3} & \textbf{5} & \textbf{10} & 14    & 18    & 24 \\
          & NWHFC & \textbf{3} & 4     & 5     & 7     & 6     & 3 \\
    35 dB & HySime & \textbf{3} & \textbf{5} & \textbf{10} & 14    & 18    & 24 \\
          & GENE-CH & \textbf{3} & \textbf{5} & \textbf{10} & \textbf{15} & 19    & 27 \\
          & GENE-AH & \textbf{3} & \textbf{5} & \textbf{10} & \textbf{15} & 18    & 24 \\
					\midrule
          & HySURE & \textbf{3} & \textbf{5} & 9     & 13    & 14    & 18 \\
          & NWHFC & \textbf{3} & 4     & 5     & 6     & 4     & 3 \\
    25 dB & HySime & \textbf{3} & \textbf{5} & 9     & 13    & 13    & 16 \\
          & GENE-CH & \textbf{3} & \textbf{5} & \textbf{10} & \textbf{15} & 16    & 21 \\
          & GENE-AH & \textbf{3} & \textbf{5} & \textbf{10} & 14    & 12    & 17 \\
					\midrule
          & HySURE & \textbf{3} & \textbf{5} & 9     & 12    & 11    & 14 \\
          & NWHFC & \textbf{3} & 4     & 4     & 5     & 2     & 3 \\
    20 dB & HySime & \textbf{3} & \textbf{5} & 9     & 12    & 9     & 12 \\
          & GENE-CH & \textbf{3} & \textbf{5} & 9     & 13    & 13    & 16 \\
          & GENE-AH & \textbf{3} & \textbf{5} & 9     & 11    & 9     & 13 \\
					\midrule
          & HySURE & \textbf{3} & \textbf{5} & 8     & 9     & 9     & 10 \\
          & NWHFC & \textbf{3} & 3     & 4     & 3     & 1     & 2 \\
    15 dB & HySime & \textbf{3} & \textbf{5} & 6     & 9     & 7     & 9 \\
          & GENE-CH & \textbf{3} & \textbf{5} & 7     & 12    & 9     & 12 \\
          & GENE-AH & \textbf{3} & \textbf{5} & 7     & 8     & 8     & 9 \\
					\midrule
          & HySURE & \textbf{3} & \textbf{5} & 6     & 8     & 7     & 7 \\
          & NWHFC & \textbf{3} & 3     & 3     & 2     & 1     & 2 \\
     10 dB & HySime & \textbf{3} & \textbf{5} & 5     & 7     & 6     & 6 \\
          & GENE-CH & \textbf{3} & 4     & 6     & 10    & 8     & 8 \\
          & GENE-AH & \textbf{3} & 4     & 5     & 7     & 7     & 7 \\

    \bottomrule
    \end{tabular}%
  \label{tab:rank_sel_syn_Data2_E2}%
\end{table}%

\section{REAL DATA EXPERIMENTS}
\label{sec:expReal}

In real data experiments, we use two data sets, Indian Pines \cite{MultiSpec} and Cuprite \cite{Cuprite}. 
Indian Pines is a widely used hyperspectral data set captured by the airborne hyperspectral sensor AVIRIS (Airborne Visible/Infrared Imaging Spectrometer) with 20 m spatial resolution per pixel and 10 nm spectral resolution per band. The data is composed of 145$\times$145 pixels in 220 bands. Cuprite is also an AVIRIS data set composed of 512$\times$614 pixels in 224 spectral bands with 10 nm spectral resolution per band. Here, Indian Pines is used for both model selection and rank selection and Cuprite is only used for the rank selection experiment.

\subsection{Hyperspectral Model Selection Using SURE}
\label{subsec: HSIMSReal}

 Fig. \ref{fig:Real_sure} compares the SURE values for models (1-5) as a function of the tuning parameter, $\lambda$. It can be seen that, SURE selects $\lambda=0$ for model (3), and model (5) has the lowest SURE value for Indian Pines compared to the other four full-rank models. Among the wavelet based models (1-3) the spectral-spatial wavelet-based model (1) outperforms the spatial model (2) and the spectral model (3).

Also from Fig. \ref{fig:Real_sure}, it can be seen that model (4), in which the unknown signal is modeled by using the spectral eigenvectors, and model (5), where a spatial wavelet is used with the spectral eigenvectors, substantially outperform wavelet-based models (1-3). 
\begin{figure}[t]
	\centering
		\includegraphics[width=8.5cm]{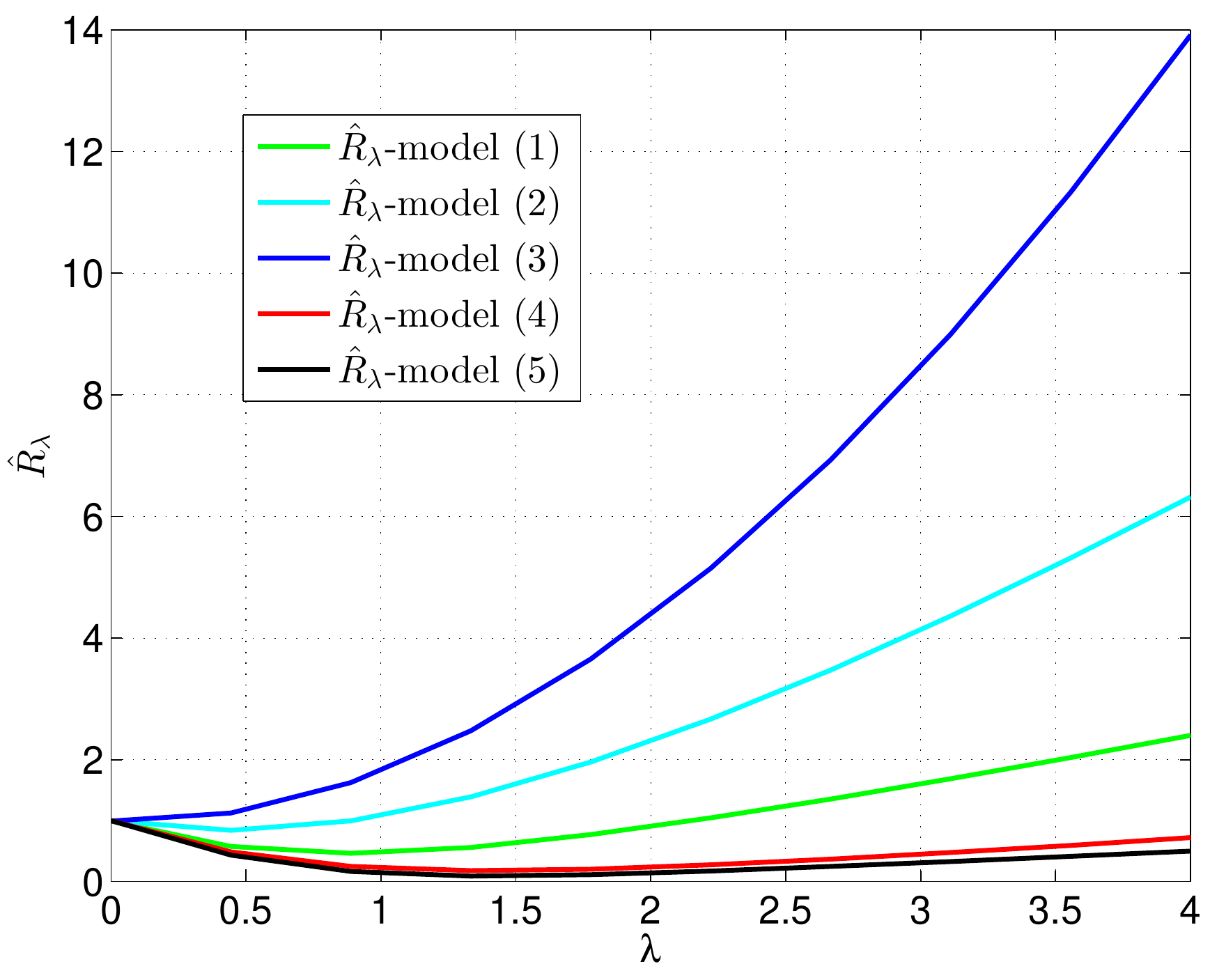}
		\caption{A comparison of SURE for models (1-5) w.r.t. the tuning parameter $\lambda$ for Indian Pines.}
	\label{fig:Real_sure}
\end{figure}

Fig. \ref{fig:Real_sure_lr_ps} compares SURE values for the low-rank models (6) and (7) at their optimum rank w.r.t. the tuning parameter, $\lambda$. It can be seen that, model (7) outperforms model (6) based on SURE values making it the best model. Also, the SURE values for the full-rank models (4) and (5) are given which clearly show the advantages of the low-rank models. Therefore, based on the results shown, model (7) is the best model among the candidates.

\begin{figure}[t]
	\centering
		\includegraphics[width=8.5cm]{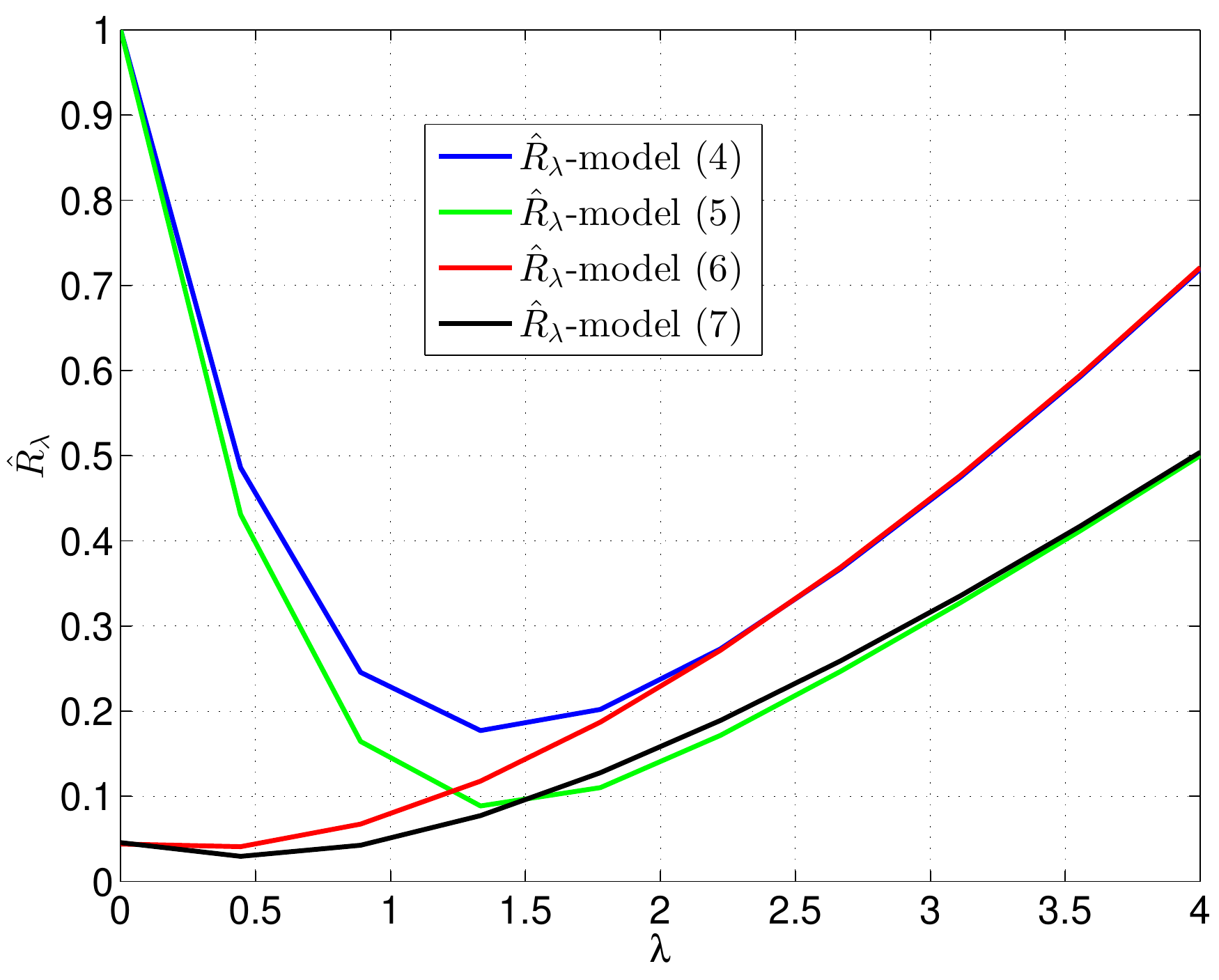}
		\caption{A comparison of SURE for models (4-7) w.r.t. the tuning parameter $\lambda$ for Indian Pines.}
	\label{fig:Real_sure_lr_ps}
\end{figure}

\subsection{Hyperspectral Rank Selection Using HySURE}
\label{subsec: HRSReal}
Here, a similar experiment is carried out on real hyperspectral data, Indian Pines, and the performance of HySURE is shown by a contour plot in Fig. \ref{fig:Contour_m7}. In Fig. \ref{fig:Contour_m7}, a logarithmic scale of the SURE value is shown for $1\leq r\leq 224$ and $0\leq\lambda\leq 4$. The minimum SURE value, as indicated in the figure, happens for $r=21$ and $\lambda=0.44$ and therefore HySURE selects $r=21$ for Indian Pines.


%


\begin{figure}[htbp]
	\centering
		\includegraphics[width=8.5cm]{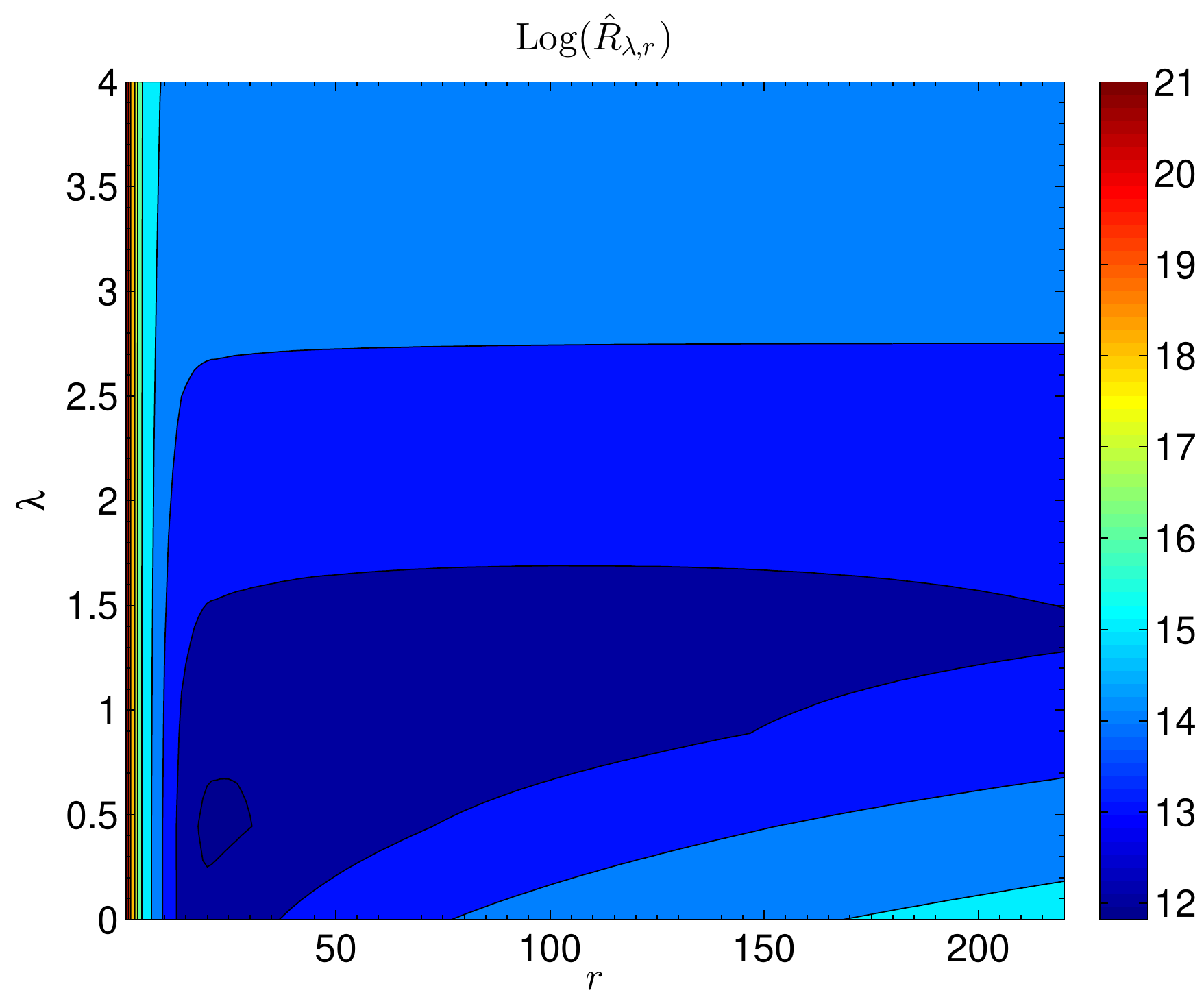}
		\caption{The contour plot of a logarithmic scale of HySURE	 w.r.t. the rank number, $r$, and the sparsity tuning parameter, $\lambda$, for Indian Pines.}
	\label{fig:Contour_m7}
\end{figure}


\subsubsection{Indian Pines Rank Estimation}

Here, we compare HySURE with HySime, GENE-AH and GENE-CH in terms of rank selection for Indian Pines data set. Usually, hyperspectral data sets contain noisy and water absorption bands. These bands are highly corrupted and usually affect the HSI analysis techniques which are not robust to the noise power and therefore they are usually discarded before HSI analysis. For instance, in \cite{HySime} and \cite{GENE} these bands have been removed before estimating the HSI rank in real experiments. Recently, it was shown that recovering those corrupted bands can improve HSI analysis \cite{RastiPhDThesis}. As we have shown in the previous section HySURE is highly robust to the noise power since the signal estimation is done based on a sparsity technique. As a result, we consider rank selection methods in two different cases. First, we apply the techniques on the data set containing all available bands which are 220 in the case of Indian Pines. Second, we will also consider the results of rank selection techniques when the noisy and water absorption bands are removed from the data sets which leads to 186 bands for Indian Pines (removed band numbers include 1-4, 103-113, 148-166). Usually, 16 ground truth materials are considered for the Indian Pines \cite{MultiSpec}. The rank estimation results are given in Table \ref{tab:ranlSelReal}. In this experiment, $P_{FA}$ is set to $10^{-8}$ for NWHFC, GENE-CH and GENE-AH and $r_{max}$ is set to 100 and 35 for GENE-CH and GENE-AH, respectively.

\subsubsection{Cuprite Rank Estimation}

Due to the natural characteristics of Cuprite Nevada scene, Cuprite data set has been widely used for applications such as rank and endmember estimation and unmixing. Therefore, we also use this AVIRIS data set for rank estimation in this subsection. Usually, 25 ground truth materials are considered for the Cuprite data set \cite{CupriteMap}. The rank estimation results are given in Table \ref{tab:ranlSelReal} for HySURE, NWHFC, HySime, GENE-CH and GENE-AH, they are applied on data having all the available bands (206 bands) and also on data set having 188 bands (after removing noisy and water absorption bands which includes bands 1-2, 104-113, 148-167 and 221-224). In this experiment, $r_{max}$ is set to 100 and 35 for GENE-CH and GENE-AH, respectively. 

\subsubsection{Discussion}

It is hard to state the true rank in the case of real hyperspectral data but we can claim that the minimum rank number is the number of classes. Therefore, in Table \ref{tab:ranlSelReal} we can see that HySime underestimates and GENE-CH overestimates the rank number for both data sets. In the case of Indian Pines, when the data set contains the corrupted bands NWHFC estimates $r=16$ which is the number of classes and after removing those bands NWHFC estimates $r=17$ which is one more than the number of classes. In the case of Cuprite, NWHFC underestimates the rank number. The estimated ranks by GENE-AH and HySURE agree quite well for all cases and they are estimated by a small difference over the number of classes. Note that GENE-CH fails to estimate a rank number in the case of Cuprite when corrupted bands are left in the data set where shown by zero in the table.



\begin{table}[htbp]
  \addtolength{\tabcolsep}{4pt}
	\centering
  \caption{Hyperspectral rank selection by applying HySime, GENEH-CH, GENE-AH, SPAMARS and SPAWMARS using all the bands of Indian Pines and Cuprite data sets and also when noisy and water absorption bands are removed (left and right column, respectively).}
 \begin{tabular}{||l||c|c||c|c||}
		\hline\hline
    & \multicolumn{2}{c||}{Indian Pines} & \multicolumn{2}{c||}{Cuprite}   \\ \hline\hline
Number of bands		& 220 & 186 & 206 &188 \\ \hline\hline
HySURE	   & 21 & 20&29&29 \\ \hline
NWHFC	   & 16 &17 &22 &17 \\ \hline
HySime  & 12 & 14&3&15\\ \hline
GENE-CH	   & 49 & 63& 0 &85\\ \hline
GENE-AH	   & 20 & 19&28&28\\ \hline

		\hline
    \end{tabular}%
  \label{tab:ranlSelReal}%
\end{table}%

\section{conclusion}
\label{sec:con}


In this paper, SURE was proposed as the rank number and the sparsity tuning parameter selection method for hyperspectral data. It was shown that SURE can simultaneously select the rank and sparsity tuning parameters for hyperspectral data.

 Based on SURE and a sparse model estimation, a technique was developed for HSI rank selection called HySURE. The performance of HySURE was evaluated by using both simulated and real hyperspectral data sets. Additionally, SURE was suggested for model evaluation in the case of real hyperspectral data and it was shown that low-rank models outperform the full-rank models.

In experiments, HySURE showed robust performances both w.r.t. the noise power and the rank number for the simulated hyperspectral data sets compared to the other rank selection techniques used in the paper. The developed method was also used to select the rank of two real data set, Indian Pines and Cuprite which gave reasonable rank numbers compared to the number of ground truth material in both cases. The method was also shown to be robust even when the noisy and water absorption bands were left in the real data sets.

%

The developed method, HySURE, is simple, automatic and easy to apply on a real HSI. 

\appendices
\section{Derivation of the shrinkage function}
\label{App: Shrinkage}

Here, we prove that the shrinkage function is the solution of the following convex minimization
\begin{equation*}
	\arg\min_{\bf {W}_r}~\frac{1}{2}\left\|{\bf Y}-{\bf AW}_r{\bf M}_r^T\right\|^{2}_{F}+\sum_{t,k}\lambda\left|w_{tk}\right|,
\end{equation*}
where ${\bf M}_r$ and ${\bf A}$ are orthogonal. The fidelity term of the minimization problem can be rewritten as
\begin{align*}
	&\left\|{\bf Y}-{\bf A}{\bf W}_r{\bf M}_r^T\right\|^{2}_{F}=\\ \nonumber
	&\mbox{tr}\left(\left({\bf Y}-{\bf AW}_r{\bf M}_r^T\right)^T\left({\bf Y}-{\bf AW}_r{\bf M}_r^T\right)\right)=\\ \nonumber
	&\mbox{tr}\left({\bf Y}^T{\bf Y}\right)-2\mbox{tr}\left({\bf M}_r{\bf W}_r^T{\bf A}^T{\bf Y}\right)+\mbox{tr}\left({\bf M}_r{\bf W}_r^T{\bf A}^T{\bf A}{\bf W}_r{\bf M}_r^T\right)
\end{align*}
where ${\bf A}^T{\bf A}={\bf I}_n$ and ${\bf M}_r^T{\bf M}_r={\bf I}_{r}$ (${\bf M}_r$ and ${\bf A}$ are orthogonal). Thus, by using trace properties and ignoring irrelevant terms, the minimization problem is given by
\begin{equation}\label{eq:MinProb}
	\arg\min_{\bf {W}_r}~\frac{1}{2}\mbox{tr}\left({\bf W}_r^T{\bf W}_r\right)-\mbox{tr}\left({\bf B}{\bf W}_r^T\right)+\sum_{t,k}\lambda\left|w_{tk}\right|,
\end{equation}
where ${\bf B}={\bf A}^T{\bf Y}{\bf M}_r=\left[b_{tk}\right]$. By adding the quadratic term $\frac{1}{2}\mbox{tr}\left({\bf B}^T{\bf B}\right)$ (which is a constant), the minimization problem (\ref{eq:MinProb}) can be written as
\begin{equation*}
	\arg\min_{\bf {W}_r}~\frac{1}{2}\left\|{\bf W}_r-{\bf B}\right\|_F^2+\sum_{t,k}\lambda\left|w_{tk}\right|,
\end{equation*}
which is a separable problem and can be solved pixelwise as
\begin{equation}\label{eq:ell_1}
	\arg\min_{{w}_{tk}}~\frac{1}{2}\left({ w}_{tk}-{b}_{tk}\right)^2+\lambda\left|w_{tk}\right|.
\end{equation}
Finally, it can be shown that the solution to (\ref{eq:ell_1}) is given by the shrinkage function
\begin{equation*}
\hat{w}_{tk}=\max\left(0,\left|b_{tk}\right|-\lambda\right) \frac{b_{tk}}{\left|b_{tk}\right|}.
\end{equation*}

%

\section{Derivation of HySURE}
\label{App: SURE}
Here, we derive a SURE formula for the sparse estimation and the general model
\begin{equation*}
{\bf Y} = {\bf A}{\bf W}_r{\bf M}_r^T + {\bf N},
\end{equation*}
where  ${\bf A}=\left[a_{it}\right]$, ${\bf M}=\left[m_{jk}\right]$ and ${\bf W}_r=\left[w_{tk}\right]$. Assuming ${\bf X}={\bf A}{\bf F}=\left[ x_{ij}\right]$ and ${\bf F}={\bf W}_r{\bf M}_r^T=\left[f_{tj}\right]$ then SURE is given by \cite{Magnus}
\begin{equation*}
{\hat R}_{{ \lambda},r} =\left\|{\bf E}\right\|_F^2+2\sum_{j=1}^p\sigma_j^2\mbox{tr}\left(\frac{\partial{\hat{\bf x}}_{(j)}}{\partial{\bf y}_{(j)}^T}\right)-np,
\end{equation*}
where ${\bf E}={\bf Y}-{\bf A}{\hat{\bf W}}_r{\bf M}_r^T$ is the residual and $\sigma_j=1$ due to prewhitening. Equivalently, we can write
\begin{equation*}
{\hat R}_{{ \lambda},r}=\left\|{\bf E}\right\|_F^2+2\sum_{i=1}^n\sum_{j=1}^p\frac{\partial{\hat{x}}_{ij}}{\partial{y}_{ij}}-np.
\end{equation*}
Also, we assume ${\bf B}={\bf A}^T{\bf Q}=\left[b_{tk}\right]$ and ${\bf Q}={\bf Y}{\bf M}_r=\left[q_{ik}\right]$. By using the chain rule we have
\begin{equation}\label{eq: partial}
\frac{\partial{\hat{x}}_{ij}}{\partial{y}_{ij}}=\frac{\partial{\hat{x}}_{ij}}{\partial{\hat{f}}_{tj}}\frac{\partial{\hat{f}}_{tj}}{\partial{\hat{w}}_{tk}}\frac{\partial{\hat{w}}_{tk}}{\partial{b}_{tk}}\frac{\partial{b}_{tk}}{\partial{q}_{ik}}\frac{\partial{q}_{ik}}{\partial{y}_{ij}},
\end{equation}
where the partial derivatives are given by
\begin{align*}
&\frac{\partial{\hat{x}}_{ij}}{\partial{\hat{f}}_{tj}}=\frac{\partial}{\partial{\hat{f}}_{tj}}\sum_{t=1}^{n}a_{it}{\hat{f}}_{tj}=\sum_{t=1}^{n}a_{it},\\
&\frac{\partial{\hat{f}}_{tj}}{\partial{\hat{w}}_{tk}}=\frac{\partial}{\partial{\hat{w}}_{tk}}\sum_{k=1}^{r}{\hat w}_{tk}m_{jk}=\sum_{k=1}^{r}m_{jk},\\
&\frac{\partial{\hat{w}}_{tk}}{\partial{b}_{tk}}={I}(\left|b_{tk}\right|>\lambda),\\
&\frac{\partial{b}_{tk}}{\partial{q}_{ik}}=\frac{\partial}{\partial{q}_{ik}}\sum_{i=1}^{n}a_{it}q_{ik}=\sum_{i=1}^{n}a_{it},\\
&\frac{\partial{q}_{ik}}{\partial{y}_{ij}}=\frac{\partial}{\partial{y}_{ij}}\sum_{j=1}^{p}y_{ij}m_{jk}=\sum_{j=1}^{p}m_{jk}.
\end{align*}
By substituting the partial derivatives in (\ref{eq: partial}), SURE is given by
\begin{align*}
&{\hat R}_{{ \lambda},r}=\left\|{\bf E}\right\|_F^2+\\
&2\sum_{i=1}^n\sum_{j=1}^p\sum_{t=1}^{n}a_{it}\sum_{k=1}^{r}m_{jk}{I}(\left|b_{tk}\right|>\lambda)\sum_{i=1}^{n}a_{it}\sum_{j=1}^{p}m_{jk}-np\\
&=\left\|{\bf E}\right\|_F^2+2\sum_{t=1}^{n}\sum_{i=1}^{n}a^2_{it}\sum_{k=1}^{r}\sum_{j=1}^{p}m^2_{jk}{I}(\left|b_{tk}\right|>\lambda)-np.
\end{align*}
Since ${\bf A}^T{\bf A}={\bf I}_n$ and ${\bf M}_r^T{\bf M}_r={\bf I}_{r}$ then we have
\begin{equation*}
	 \sum_{t=1}^{n}\sum_{i=1}^{n}a^2_{it}=1~~~\mbox{and}~~~\sum_{k=1}^{r}\sum_{j=1}^{p}m^2_{jk}=1.
\end{equation*}
Finally, SURE is given by
\begin{equation}\label{eq: SUREgen}
{\hat R}_{{ \lambda},r}= \left\|{\bf E}\right\|_F^2+2\sum_{t=1}^n\sum_{k=1}^r{I}(\left|b_{tk}\right|>\lambda)-np.
\end{equation}
The tuning parameters $\lambda$ and $r$ and the models are selected corresponding to the minimum of SURE. 
It is worth to mention that based on the derivation above, the solution is also true for the case when ${\bf A}$ and ${\bf M}_r$ are low-rank orthogonal matrices.


\section*{Acknowledgment}
This work was supported by the University of Iceland
Research Fund, and the Icelandic research fund (130635051).
\bibliographystyle{IEEEtran}
\bibliography{IEEEabrv,refs}

\begin{thebibliography}{10}
\providecommand{\url}[1]{#1}
\csname url@samestyle\endcsname
\providecommand{\newblock}{\relax}
\providecommand{\bibinfo}[2]{#2}
\providecommand{\BIBentrySTDinterwordspacing}{\spaceskip=0pt\relax}
\providecommand{\BIBentryALTinterwordstretchfactor}{4}
\providecommand{\BIBentryALTinterwordspacing}{\spaceskip=\fontdimen2\font plus
\BIBentryALTinterwordstretchfactor\fontdimen3\font minus
  \fontdimen4\font\relax}
\providecommand{\BIBforeignlanguage}[2]{{%
\expandafter\ifx\csname l@#1\endcsname\relax
\typeout{** WARNING: IEEEtran.bst: No hyphenation pattern has been}%
\typeout{** loaded for the language `#1'. Using the pattern for}%
\typeout{** the default language instead.}%
\else
\language=\csname l@#1\endcsname
\fi
#2}}
\providecommand{\BIBdecl}{\relax}
\BIBdecl

\bibitem{varshney2010advanced}
P.~Varshney and M.~Arora, \emph{Advanced Image Processing Techniques for
  Remotely Sensed Hyperspectral Data}.\hskip 1em plus 0.5em minus 0.4em\relax
  Springer, 2010.

\bibitem{RastiSPIE13UWT}
B.~Rasti, J.~R. Sveinsson, M.~O. Ulfarsson, and J.~A. Benediktsson,
  ``Hyperspectral image restoration using wavelets,'' \emph{Proc. SPIE}, vol.
  8892, pp. 889\,207--889\,207--9, 2013.

\bibitem{WSRRR}
B.~Rasti, J.~Sveinsson, and M.~Ulfarsson, ``Wavelet-based sparse reduced-rank
  regression for hyperspectral image restoration,'' \emph{IEEE Transactions on
  Geoscience and Remote Sensing}, vol.~52, no.~10, pp. 6688--6698, {O}ct 2014.

\bibitem{BourennaneTD}
N.~Renard and S.~Bourennane, ``Improvement of target detection methods by
  multiway filtering,'' \emph{IEEE Transactions on Geoscience and Remote
  Sensing}, vol.~46, no.~8, pp. 2407--2417, {A}ug. 2008.

\bibitem{HySime}
J.~Bioucas-Dias and J.~Nascimento, ``Hyperspectral subspace identification,''
  \emph{IEEE Transactions on Geoscience and Remote Sensing}, vol.~46, no.~8,
  pp. 2435--2445, 2008.

\bibitem{Keshava2003A}
N.~Keshava, ``{A Survey of Spectral Unmixing Algorithms},'' \emph{Lincoln
  Laboratory Journal}, vol.~14, no.~1, pp. 55--78, 2003.

\bibitem{Changsubs}
C.-I. Chang and Q.~Du, ``{Estimation of number of spectrally distinct signal
  sources in hyperspectral imagery},'' \emph{IEEE Transactions on Geoscience
  and Remote Sensing}, vol.~42, no.~3, pp. 608--619, {M}ar. 2004.

\bibitem{HFC}
J.~Harsanyi, W.~Farrand, and C.~Chang, ``Determining the number and identity of
  spectral endmembers: An integrated approach using {N}eyman-{P}earson
  eigenthresholding and iterative constrained {RMS} error minimization,'' in
  \emph{9th Thematic Conf. Geologic Remote Sens.}, 1993.

\bibitem{AIC_MDL}
C.-I. Chang and Q.~Du, ``Estimation of number of spectrally distinct signal
  sources in hyperspectral imagery,'' \emph{IEEE Transactions on Geoscience and
  Remote Sensing}, vol.~42, no.~3, pp. 608--619, {M}arch 2004.

\bibitem{MDL}
J.~Rissanen, ``Modeling by shortest data description,'' \emph{Automatica},
  vol.~14, no.~5, pp. 465 -- 471, 1978.

\bibitem{AIC}
H.~Akaike, ``A new look at the statistical model identification,'' \emph{IEEE
  Transactions on Automatic Control}, vol.~19, no.~6, pp. 716--723, {D}ec 1974.

\bibitem{ELM}
B.~Luo, J.~Chanussot, S.~Doute, and L.~Zhang, ``Empirical automatic estimation
  of the number of endmembers in hyperspectral images,'' \emph{IEEE Geoscience
  and Remote Sensing Letters}, vol.~10, no.~1, pp. 24--28, {J}an. 2013.

\bibitem{Zare2007_RS}
A.~Zare and P.~Gader, ``Sparsity promoting iterated constrained endmember
  detection in hyperspectral imagery,'' \emph{IEEE Geoscience and Remote
  Sensing Letters}, vol.~4, no.~3, pp. 446--450, {J}uly 2007.

\bibitem{ICE}
M.~Berman, H.~Kiiveri, R.~Lagerstrom, A.~Ernst, R.~Dunne, and J.~Huntington,
  ``{ICE}: a statistical approach to identifying endmembers in hyperspectral
  images,'' \emph{IEEE Transactions on Geoscience and Remote Sensing}, vol.~42,
  no.~10, pp. 2085--2095, {O}ct. 2004.

\bibitem{rarevectors}
O.~Kuybeda, D.~Malah, and M.~Barzohar, ``Rank estimation and redundancy
  reduction of high-dimensional noisy signals with preservation of rare
  vectors,'' \emph{IEEE Transactions on Signal Processing}, vol.~55, no.~12,
  pp. 5579--5592, {D}ec 2007.

\bibitem{MSE_Rank_sel}
M.~Farzam and S.~Beheshti, ``Simultaneous denoising and intrinsic order
  selection in hyperspectral imaging,'' \emph{IEEE Transactions on Geoscience
  and Remote Sensing}, vol.~49, no.~9, pp. 3423--3436, {S}ep. 2011.

\bibitem{GENE}
A.~Ambikapathi, T.-H. Chan, C.-Y. Chi, and K.~Keizer, ``Hyperspectral data
  geometry-based estimation of number of endmembers using p-norm-based pure
  pixel identification algorithm,'' \emph{IEEE Transactions on Geoscience and
  Remote Sensing}, vol.~51, no.~5, pp. 2753--2769, {M}ay 2013.

\bibitem{RastiB}
B.~Rasti, J.~R. Sveinsson, M.~O. Ulfarsson, and J.~A. Benediktsson,
  ``Hyperspectral image denoising using 3{D} wavelets,'' in \emph{Proceedings
  of International Geoscience and Remote Sensing Symposium (IGARSS)}, {J}uly
  2012, pp. 1349--1352.

\bibitem{EladBook}
M.~Elad, \emph{Sparse and Redundant Representations: From Theory to
  Applications in Signal and Image Processing}, 1st~ed.\hskip 1em plus 0.5em
  minus 0.4em\relax Springer Publishing Company, Incorporated, 2010.

\bibitem{Vetterli1995}
M.~Vetterli and J.~Kovacevic, \emph{{Wavelets and Subband Coding (Prentice Hall
  Signal Processing Series)}}.\hskip 1em plus 0.5em minus 0.4em\relax Prentice
  Hall PTR, 1995.

\bibitem{Strang}
G.~Strang and T.~Nguyen, \emph{Wavelets and Filter Banks}, 1st~ed.\hskip 1em
  plus 0.5em minus 0.4em\relax Wellesley-Cambridge, 1996.

\bibitem{MP}
S.~Mallat and Z.~Zhang, ``Matching pursuits with time-frequency dictionaries,''
  \emph{IEEE Transactions on Signal Processing}, vol.~41, no.~12, pp.
  3397--3415, {D}ec. 1993.

\bibitem{Tibshirani94}
R.~Tibshirani, ``Regression shrinkage and selection via the lasso,''
  \emph{Journal of the Royal Statistical Society, Series B}, vol.~58, pp.
  267--288, 1994.

\bibitem{Chen98atomicdecomposition}
S.~Chen, D.~Donoho, and M.~Saunders, ``Atomic decomposition by basis pursuit,''
  \emph{SIAM Journal on Scientific Computing}, vol.~20, pp. 33--61, 1998.

\bibitem{FORPjour}
B.~Rasti, J.~R. Sveinsson, M.~O. Ulfarsson, and J.~A. Benediktsson,
  ``Hyperspectral image denoising using first order spectral roughness penalty
  in wavelet domain,'' \emph{IEEE Journal of Selected Topics in Applied Earth
  Observations and Remote Sensing}, vol.~7, no.~6, pp. 2458--2467, {J}une 2014.

\bibitem{hypeGMCA}
Y.~Moudden and J.~Bobin, ``Hyperspectral {BSS} using {GMCA} with
  {S}patio-{S}pectral {S}parsity {C}onstraints,'' \emph{IEEE Transactions on
  Image Processing}, vol.~20, no.~3, pp. 872--879, 2011.

\bibitem{HyInpaint}
\BIBentryALTinterwordspacing
 [Online]. Available:
  \url{http://md.cosmostat.org/Hyperspectral_Data_-_Inpainting.html}
\BIBentrySTDinterwordspacing

\bibitem{CVMS}
S.~Arlot and A.~Celisse, ``{A survey of cross-validation procedures for model
  selection},'' \emph{Statistics Surveys}, vol.~4, no.~0, pp. 40--79, 2010.

\bibitem{modelCV}
J.~Shao, ``Linear model selection by cross-validation,'' \emph{Journal of the
  American Statistical Association}, vol.~88, no. 422, pp. 486--494, {J}un.
  1993.

\bibitem{CVSHURE}
S.~Ramani, Z.~Liu, J.~Rosen, J.-F. Nielsen, and J.~Fessler, ``Regularization
  parameter selection for nonlinear iterative image restoration and {MRI}
  reconstruction using {GCV} and {SURE}-based methods,'' \emph{IEEE
  Transactions on Image Processing}, vol.~21, no.~8, pp. 3659--3672, {A}ug.
  2012.

\bibitem{CVTV}
H.~Liao, F.~Li, and M.~K. Ng, ``Selection of regularization parameter in total
  variation image restoration,'' \emph{J. Opt. Soc. Am. A}, vol.~26, pp.
  2311--2320, {O}ct. 2009.

\bibitem{CVTVRS}
P.~Liu and D.~Liu, ``Selection of regularization parameter based on generalized
  cross-validation in total variation remote sensing image restoration,'' in
  \emph{Proc. SPIE}, vol. 7830, 2010, pp. 78\,301P--78\,301P--7.

\bibitem{SURE}
C.~M. Stein, ``Estimation of the mean of a multivariate normal distribution,''
  \emph{The Annals of Statistics}, vol.~9, no.~6, pp. 1135--1151, Nov.,1981.

\bibitem{Donoho95adaptingto}
D.~Donoho and I.~Johnstone, ``Adapting to unknown smoothness via wavelet
  shrinkage,'' \emph{Journal of the American Statistical Association}, vol.~90,
  pp. 1200--1224, 1995.

\bibitem{SoloSURE}
V.~Solo, ``A {SURE}-fired way to choose smoothing parameters in ill-conditioned
  inverse problems,'' in \emph{International Conference on Image Processing
  (ICIP)}, vol.~3, 1996, pp. 89--92.

\bibitem{Eldar}
Y.~Eldar, ``Generalized {SURE} for exponential families: Applications to
  regularization,'' \emph{IEEE Transactions on Signal Processing}, vol.~57,
  no.~2, pp. 471--481, {F}eb. 2009.

\bibitem{mou}
M.~Ulfarsson and V.~Solo, ``Tuning parameter selection for underdetermined
  reduced-rank regression,'' \emph{IEEE Signal Processing Letters}, vol.~20,
  no.~9, pp. 881--884, Sept 2013.

\bibitem{Magnus}
------, ``Dimension estimation in noisy {PCA} with {SURE} and random matrix
  theory,'' \emph{IEEE Transactions on Signal Processing}, vol.~56, no.~12, pp.
  5804 --5816, {D}ec. 2008.

\bibitem{USGS}
\BIBentryALTinterwordspacing
D.~R.~N. Clark, ``{USGS} digital spectral library,'' {S}eptember 2007.
  [Online]. Available: \url{http://speclab.cr.usgs.gov/spectral-lib.html}
\BIBentrySTDinterwordspacing

\bibitem{HySimeCode}
\BIBentryALTinterwordspacing
 [Online]. Available: \url{http://www.lx.it.pt/~bioucas/code.htm}
\BIBentrySTDinterwordspacing

\bibitem{MultiSpec}
\BIBentryALTinterwordspacing
 [Online]. Available:
  \url{https://engineering.purdue.edu/~biehl/MultiSpec/hyperspectral.html}
\BIBentrySTDinterwordspacing

\bibitem{Cuprite}
\BIBentryALTinterwordspacing
 [Online]. Available: \url{http://aviris.jpl.nasa.gov/data/free\_data.html}
\BIBentrySTDinterwordspacing

\bibitem{RastiPhDThesis}
B.~Rasti, ``{Sparse Hyperspectral Image Modeling and Restoration},'' Ph.D.
  dissertation, University of Iceland, {D}ec. 2014.

\bibitem{CupriteMap}
\BIBentryALTinterwordspacing
 [Online]. Available:
  \url{http://speclab.cr.usgs.gov/cuprite95.tgif.2.2um\_map.gif}
\BIBentrySTDinterwordspacing

\end{thebibliography}
\end{document}